\documentclass[runningheads]{svproc}
\usepackage{amsmath}
\usepackage{amssymb}
\usepackage{mathtools}
\usepackage{amsfonts}
\usepackage{mathtools}
\usepackage{graphicx}
\usepackage{multirow}
\usepackage{subfig}
\usepackage{footnote}

\usepackage{hyperref}
\usepackage{xcolor}
\usepackage{float}
\newcommand{\eg}{e.\,g., }
\newcommand{\ie}{i.\,e., }

\begin{document}
\title{TimeGNN: Temporal Dynamic Graph Learning for Time Series Forecasting}
%
%
\author{Nancy Xu\inst{1} \and
Chrysoula Kosma\inst{2} \and
Michalis Vazirgiannis\inst{1,2}}
\authorrunning{N. Xu et al.}
%
\institute{KTH Royal Institute of Technology, Stockholm, Sweden \\
\email{nancyx@kth.se}\and
\'Ecole Polytechnique IPP, Palaiseau, France \\
\email{\{kosma,mvazirg\}@lix.polytechnique.fr}}
\maketitle              
\begin{abstract}
Time series forecasting lies at the core of important real-world applications in many fields of science and engineering. 
The abundance of large time series datasets that consist of complex patterns and long-term dependencies has led to the development of various neural network architectures.
Graph neural network approaches, which jointly learn a graph structure based on the correlation of raw values of multivariate time series while forecasting, have recently seen great success.
However, such solutions are often costly to train and difficult to scale. 
In this paper, we propose TimeGNN, a method that learns dynamic temporal graph representations that can capture the evolution of inter-series patterns along with the correlations of multiple series. 
TimeGNN achieves inference times 4 to 80 times faster than other state-of-the-art graph-based methods while achieving comparable forecasting performance. 

\end{abstract}

\section{Introduction}
From financial investment and market analysis \cite{ding2015deep} to traffic \cite{li2017diffusion}, electricity management, healthcare \cite{chauhan2015anomaly}, and climate science, accurately predicting the future real values of series based on available historical records forms a coveted task over time in various scientific and industrial fields. 
There are a wide variety of methods employed for time series forecasting, ranging from statistical \cite{box2015time} to recent deep learning approaches \cite{lim2021time}. However, there are several major challenges present.
Real-world time series data are often subject to noisy and irregular observations, missing values, repeated patterns of variable periodicities and very long-term dependencies.
While the time series are supposed to represent continuous phenomena, the data is usually collected using sensors. Thus, observations are determined by a sampling rate with potential information loss.
On the other hand, standard sequential neural networks, such as recurrent (RNNs) \cite{rumelhart1986learning} and convolutional networks (CNNs) \cite{lecun1998gradient}, are discrete and assume regular spacing between observations. 
Several continuous analogues of such architectures that implicitly handle the time information have been proposed to address irregularly sampled missing data \cite{rubanova2019latent}.
The variable periodicities and long-term dependencies present in the data make models prone to shape and temporal distortions, overfitting and poor local minima while training with standard loss functions (\eg MSE).
Variants of DTW and MSE have been proposed to mitigate these phenomena and can increase the forecasting quality of deep neural networks \cite{le2022deep,kosma2022time}.

A novel perspective for boosting the robustness of neural networks for complex time series is to extract representative embeddings for patterns after transforming them to another representation domain, such as the spectral one.
Spectral approaches have seen much use in the text domain. 
Graph-based text mining (\ie Graph-of-Words) \cite{rousseau2013graph} can be used for capturing the relationships between the terms and building document-level representations.
It is natural, then, that such approaches might be suitable for more general sequence modeling. 
Capitalizing on the recent success of graph neural networks (GNNs) on graph structured data, a new family of algorithms jointly learns a correlation graph between interrelated time series while simultaneously performing forecasting \cite{wu2020connecting,cao2020spectral,shang2021discrete}.
The nodes in the learnable graph structure represent each individual time series and the links between them express their temporal similarities.
However, since such methods rely on series-to-series correlations, they do not explicitly represent the inter-series temporal dynamics evolution. 
Some preliminary studies have proposed simple computational methods for mapping time series to temporal graphs where each node corresponds to a time step, such as the visibility graph \cite{lacasa2008time} and the recurrence network \cite{donner2010recurrence}.

In this paper, we propose a novel neural network, \textit{TimeGNN}, that extends these previous approaches by jointly learning dynamic temporal graphs for time series forecasting on raw data. 
TimeGNN (i) extracts temporal embeddings from sliding windows of the input series using dilated convolutions of different receptive sizes, (ii) constructs a learnable graph structure, which is forward and directed, based on the similarity of the embedding vectors in each window in a differentiable way, (iii) applies standard GNN architectures to learn embeddings for each node and produces forecasts based on the representation vector of the last time step.
We evaluate the proposed architecture on various real-world datasets and compare it against several deep learning benchmarks, including graph-based approaches. 
Our results indicate that TimeGNN is significantly less costly in both inference and training while achieving comparable forecasting performance. The code implementation for this paper is available at \url{https://github.com/xun468/Time-GNN}.

\section{Related Work}
\noindent\textbf{Time series forecasting models.}
Time series forecasting has been a long-studied challenge in several application domains.
In terms of statistical methods, linear models including the autoregressive integrated moving average (ARIMA) \cite{box2015time} and its multivariate extension, the vector autoregressive model (VAR) \cite{hamilton2020time} constitute the most dominant approaches.
The need for capturing non-linear patterns and overcoming the strong assumptions for statistical methods, \eg the stationarity assumption, has led to the application of deep neural networks, initially introduced in sequential modeling, to the time series forecasting setting. 
Those models include recurrent neural networks (RNNs) \cite{rumelhart1986learning} and their improved variants for alleviating the vanishing gradient problem, namely the LSTM \cite{hochreiter1997long} and the GRU \cite{cho2014learning}. 
An alternative method for extracting long-term dependencies via large receptive fields can be achieved by leveraging stacked dilated convolutions, as proposed along with the Temporal Convolution Network (TCN) \cite{bai2018empirical}.
Bridging CNNs and LSTMs to capture both short-term local dependency patterns among variables and long-term patterns, the Long- and Short-term Time-series network (LSTNet) \cite{lai2018modeling} has been proposed.
For univariate point forecasting, the recently proposed N-BEATS model \cite{oreshkin2019n} introduces a deep neural architecture based on a deep stack of fully-connected layers with basis expansion.
Attention-based approaches have also been employed for time-series forecasting, including
Transformer \cite{vaswani2017attention} and Informer \cite{zhou2021informer}.
Finally, for efficient long-term modeling, the most recent Autoformer architecture \cite{wu2021autoformer} introduces an auto-correlation mechanism in place of self-attention, which extracts and aggregates similar sub-series based on the series periodicity. \\
\noindent   \textbf{Graph neural networks.}
Over the past few years, graph neural networks (GNNs) have been applied with great success to machine learning problems on graphs in various fields, including chemistry for drug screening \cite{kearnes2016molecular} and biology for predicting the functions of proteins modeled as graphs \cite{gligorijevic2021structure}. 
The field of GNNs has been largely dominated by the so-called message passing neural networks (MPNNs) \cite{gilmer2017neural}, where each node updates its feature vector by aggregating the feature vectors of its neighbors. 
In the case of time series data on arbitrary known graphs, \eg in traffic forecasting, several architectures that combine sequential models with GNNs have been proposed \cite{li2017diffusion,yu2017spatio,seo2018structured,zhao2019t}. \\
\noindent \textbf{Joint graph structure learning and forecasting.}
However, since spatial-temporal forecasting requires an apriori topology which does not apply in the case of most real-world time series datasets, graph structure learning has arisen as a viable solution.
Recent models perform joint graph learning and forecasting for multivariate time series data using GNNs, intending to capture temporal patterns and exploit the interdependency among time series while predicting the series' future values.
The most dominant algorithms include NRI \cite{kipf2018neural}, MTGNN \cite{wu2020connecting} and GTS \cite{shang2021discrete}, in which the graph nodes represent the individual time series and their edges represent their temporal evolution. 
MTGNN obtains the graph adjacency from the as a degree-$k$ structure from the pairwise scores of embeddings of each series in the multivariate collection, which might pose challenges to end-to-end learning.
On the other hand, NRI and GTS employ the Gumbel softmax trick \cite{jang2016categorical} to differentiably sample a discrete adjacency matrix from the edge probabilities.
Both models compute fixed-size representations of each node based on the time series, with the former dynamically producing the representations per individual window and the latter extracting global representations from the whole training series.
MTGNN combines temporal convolution with graph convolution layers, and GTS uses a Diffusion Convolutional Recurrent Neural Network (DCRNN) \cite{li2017diffusion}, where the hidden representations of nodes are diffused using graph convolutions at each step.

\section{Method}
Let $\{\mathbf{X}_{i,1:T}\}_{i=1}^m$ be a multivariate time series that consists of $m$ channels and has a length equal to $T$. 
Then, $\mathbf{X}_t \in \mathbb{R}^{m}$ represents the observed values at time step $t$.
Let also $\mathcal{G}$ denote the set of temporal dynamic graph structures that we want to infer.

Given the observed values of $\tau$ previous time steps of the time series, \ie~$\mathbf{X}_{t-\tau}, \ldots, \mathbf{X}_{t-1}$, the goal is to forecast the next $h$ time steps (\eg~$h=1$ for 1-step forecasting), \ie~$\hat{\mathbf{X}}_{t}, \hat{\mathbf{X}}_{t+1}, \ldots,$ $\hat{\mathbf{X}}_{t+h-1}$. 
These values can be obtained by the forecasting model $\mathcal{F}$ with parameters $\Phi$ and the graphs $\mathcal{G}$ as follows:

\begin{equation}
\hat{\mathbf{X}}_{t}, \hat{\mathbf{X}}_{t+1}, \ldots, \hat{\mathbf{X}}_{t+h-1} = \mathcal{F}(\mathbf{X}_{t-\tau}, \ldots, \mathbf{X}_{t-1} ; \mathcal{G} ; \Phi)
\label{eq:1}
\end{equation}

\subsection{Time Series Feature Extraction}


\begin{figure}[t]
\centering
  \includegraphics[width=0.75\textwidth]{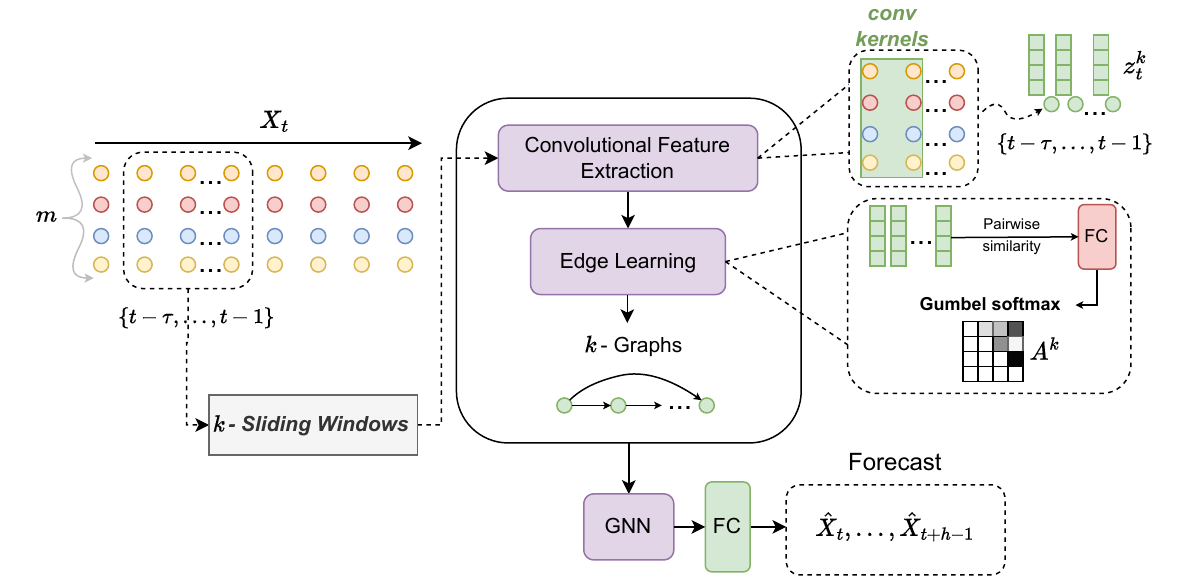}
  \caption{The proposed TimeGNN framework time series for graph learning from raw time series and forecasting based on embeddings learned on the parameterized graph structures.}
\label{fig:model}
\end{figure}


Unlike previous methods which extract one feature vector per variable in the multivariate input, our method extracts one feature vector per time step in each window $k$ of length $\tau$.
Temporal sub-patterns are learned using stacked dilated convolutions, similar to the main blocks of the inception architecture \cite{lin2013network}.

Given the sliding windows $\mathbf{S} = \{\mathbf{X}_{t-\tau+k-K}, \ldots, \mathbf{X}_{t+k-K-1}\}_{k=1}^K$, we perform the following convolutional operations to extract three feature maps $\mathbf{f}_0^k$, $\mathbf{f}_1^k$, $\mathbf{f}_2^k$, per window $\mathbf{S}^k$. 
Let $\mathbf{f}_i^k \in \mathbb{R}^{\tau \times d}$ for hidden dimension $d$ of the convolutional kernels, such that:
\begin{equation}
    \begin{split}
    &\mathbf{f}_0^k = \mathbf{S}^k \ast \mathbf{C}_0^{1,1} +\mathbf{b}_{01} \\
    &\mathbf{f}_1^k = (\mathbf{S}^k \ast \mathbf{C}_1^{1,1} +\mathbf{b}_{11}) \ast \mathbf{C}_2^{3,3} + \mathbf{b}_{23} \\
    &\mathbf{f}_2^k = (\mathbf{S}^k \ast \mathbf{C}_2^{1,1} + \mathbf{b}_{21}) \ast \mathbf{C}_2^{5,5} + \mathbf{b}_{25}
    \end{split}
\end{equation}
where $\ast$ the convolutional operator, $\mathbf{C}_0^{1,1}$, $\mathbf{C}_1^{1,1}$, $\mathbf{C}_2^{1,1}$ convolutional kernels of size 1 and dilation rate 1, $\mathbf{C}_2^{3,3}$ a convolutional kernel of size 3 and dilation rate 3, $\mathbf{C}_2^{5,5}$ a convolutional kernel of size 5 and dilation rate 5, and $\mathbf{b}_{01}, \mathbf{b}_{11}, \mathbf{b}_{21}, \mathbf{b}_{23}, \mathbf{b}_{25}$ the corresponding bias terms.

The final representations per window $k$ are obtained using a fully connected layer on the concatenated features $\mathbf{f}_0^k, \mathbf{f}_1^k, \mathbf{f}_2^k$, \ie $\mathbf{z}^k = \text{FC}(\mathbf{f}_0^k \|\mathbf{f}_1^k\|\mathbf{f}_2^k)$, such that $\mathbf{z}^k \in \mathbb{R}^{\tau \times d}$.
In the next sections, we refer to each time step of the hidden representation of the feature extraction module in each window $k$ as $\mathbf{z}_i^k, \forall~i \in \{1, \ldots \tau\}$.

\subsection{Graph Structure Learning}



The set $\mathcal{G} = \{\mathcal{G}^k\}, k \in \mathbb{N}^*$ describes the collection of graph structures that are parameterized for all individual sliding window of length $\tau$ of the series, where $K$ defines the total number of windows.
The goal of the graph learning module is to learn each adjacency matrix $\mathbf{A}^k \in \{0,1\}^{\tau \times \tau}$ for a temporal window of observations $\mathbf{S}^k$.
Following the works of \cite{kipf2018neural,shang2021discrete}, we use the Gumbel softmax trick to sample a discrete adjacency matrix as described below.

For the Gumbel softmax trick, let $\mathbf{A}^k$ refer to a random variable of the matrix Bernoulli distribution parameterized by $\boldsymbol{\theta}^k \in [0,1]^{\tau \times \tau}$, so that $A_{ij}^k \sim Ber(\theta_{ij}^k)$ is independent for pairs $(i,j)$.
By applying the Gumbel reparameterization trick \cite{jang2016categorical} for enabling differentiability in sampling, we can obtain the following:
\begin{equation}
    \begin{gathered}
    A_{ij}^k = \sigma((\log(\theta_{ij}^k/(1-\theta_{ij}^k)) + (\mathbf{g}_{i,j}^1 - \mathbf{g}_{i,j}^2))/s),\\
    \mathbf{g}_{i,j}^1,\mathbf{g}_{i,j}^2 \sim \text{Gumbel}(0,1), \forall~i,j
    \end{gathered}
\end{equation}
where $\mathbf{g}_{i,j}^1,\mathbf{g}_{i,j}^2$ are vectors of i.i.d samples drawn from Gumbel distribution, $\sigma$ is the sigmoid activation and $s$ is a parameter that controls the smoothness of samples, so that the distribution converges to categorical values as $s \xrightarrow{}0$.

The link predictor takes each pair of extracted features $(\mathbf{z}_i^k,\mathbf{z}_j^k)$ of window $k$ and maps their similarity to a $\theta_{ij}^k \in [0,1]$ by applying fully connected layers. Then the Gumbel reparameterization trick is used to approximate a sigmoid activation function while retaining differentiability:
\begin{equation}
\theta_{ij}^k = \sigma\Big( \text{FC}\big(\text{FC}(\mathbf{z}_i^k\|\mathbf{z}_j^k)\big)\Big)
\end{equation}
In order to obtain directed and forward (\ie no look-back in previous time steps in the history) graph structures $\mathcal{G}$ we only learn the upper triangular part of the adjacency matrices.

\subsection{Graph Neural Network for Forecasting}



Once the collection $\mathcal{G}$ of learnable graph structures per sliding window $k$ are sampled, standard GNN architectures can be applied for capturing the node-to-node relations, \ie the temporal graph dynamics.
GraphSAGE \cite{hamilton2017inductive} was chosen as the basic building GNN block of the node embedding learning architecture as it can effectively generalize across different graphs with the same attributes.
GraphSAGE is an inductive framework that exploits node feature information and generates node embeddings (\ie $\mathbf{h}_u$ for node $u$) via a learnable function, by sampling and aggregating features from a node's local neighborhood (\ie $\mathcal{N}(u)$). 

Let $(\mathcal{V}^k, \mathcal{E}^k)$ correspond to the set of nodes and edges of the learnable graph structure for each $\mathcal{G}^k$.
The node embedding update process for each $p \in \{1, \ldots, P\}$ aggregation steps, employs the mean-based aggregator, namely convolutional, by calculating the element-wise mean of the vectors in $\{\mathbf{h}_u^{p-1}, \forall u \in \mathcal{N}(u)\}$, such that:
\begin{equation}
\mathbf{h}_u^{p} \xleftarrow{} \sigma(\mathbf{W} \cdot \text{MEAN}(\{\mathbf{h}_u^{p-1}\} \cup \{\mathbf{h}_u^{p-1} \forall u \in \mathcal{N}(u)\}))
\end{equation}
where $\mathbf{W}$ trainable weights. 
The final normalized (\ie $\mathbf{\tilde{h}}_u^{p}$) representation of the last node (\ie time step) in each forward and directed graph denoted as $\mathbf{z}_{u_T} = \mathbf{\tilde{h}}_{u_T}^p$ is passed to the output module.
The output module consists of two fully connected layers which reduce the vector into the final output dimension, so as to correspond to the forecasts $\hat{\mathbf{X}}_{t}, \hat{\mathbf{X}}_{t+1}, \ldots, \hat{\mathbf{X}}_{t+h-1}$.
Figure \ref{fig:model} demonstrates the feature extraction, graph learning, GNN and output modules of the proposed TimeGNN architecture.
\subsection{Training \& Inference}
To train the parameters of Equation~\eqref{eq:1} for the time series point forecasting task, we use the mean absolute error loss (MAE).
Let $\mathbf{\hat{X}}^{(i)}, i \in \{1,...,K\}$ denote the predicted vector values for $K$ samples, then the MAE loss is defined as:
\[\mathcal{L} = \frac{1}{K}\sum_{i=1}^{K}\|\mathbf{\hat{X}}^{(i)}-\mathbf{X}^{(i)}\|\]


The optimized weights for the feature extraction, graph structure learning, GNN and output modules are selected based on the minimum validation loss during training, which is evaluated as described in the experimental setup (section \ref{sec:4.3})

\section{Experimental Evaluation}
We next describe the experimental setup, including the datasets and baselines used for comparisons.
We also demonstrate and analyze the results obtained by the proposed TimeGNN architecture and the baseline models.
\subsection{Datasets}
This work was evaluated on the following multivariate time series datasets: 

\noindent\textbf{Exchange-Rate} which consists of the daily exchange rates of $8$ countries from $1990$ to $2016$, following the preprocessing of \cite{lai2018modeling}.

\noindent\textbf{Weather} that contains hourly observations of $12$ climatological features over a period of four years \footnote{https://www.ncei.noaa.gov/data/local-climatological-data/}, preprocessed as in \cite{zhou2021informer}.

\noindent\textbf{Electricity-Load} is based on the UCI Electricity Consuming Load dataset \footnote{https://archive.ics.uci.edu/ml/datasets/ElectricityLoadDiagrams20112014} that records the electricity consumption of $370$ Portuguese clients from $2011$ to $2014$. As in \cite{zhou2021informer}, the recordings are binned into hourly intervals over the period of 2012 to 2014 and incomplete clients are removed. 

\noindent\textbf{Solar-Energy} contains the solar power production records in $2006$, sampled every $10$ minutes from $137$ PV plants in Alabama State \footnote{http://www.nrel.gov/grid/solar-power-data.html}.

\noindent\textbf{Traffic} is a collection of $48$ months, between $2015$ and $2016$, of hourly data from the California Department of Transportation \footnote{http://pems.dot.ca.gov}.
The data describes the road occupancy rates (between $0$ and $1$) measured by different sensors. 

\subsection{Baselines}
We consider five baseline models for comparison with our TimeGNN proposed architecture. 
We chose two graph-based methods, MTGNN\cite{wu2020connecting} and GTS\cite{shang2021discrete}, and three non graph-based methods, LSTNet\cite{lai2018modeling}, LSTM\cite{hochreiter1997long}, and TCN\cite{bai2018empirical}. 
Also, we evaluate the performance of TimeMTGNN, a variant of MTGNN that includes our proposed graph learning module.
LSTM and TCN follow the size of the hidden dimension and number of layers of TimeGNN. 
Those were fixed to three layers with hidden dimensions of $32$, $64$ for the Exchange-Rate and Weather datasets and $128$ for Electricity, Solar-Energy and Traffic.
In the case of MTGNN, GTS, and LSTNet, parameters were kept as close as possible to the ones mentioned in their experimental setups. 

\begin{figure}[t]
    \centering
    \includegraphics[width=0.85\textwidth]{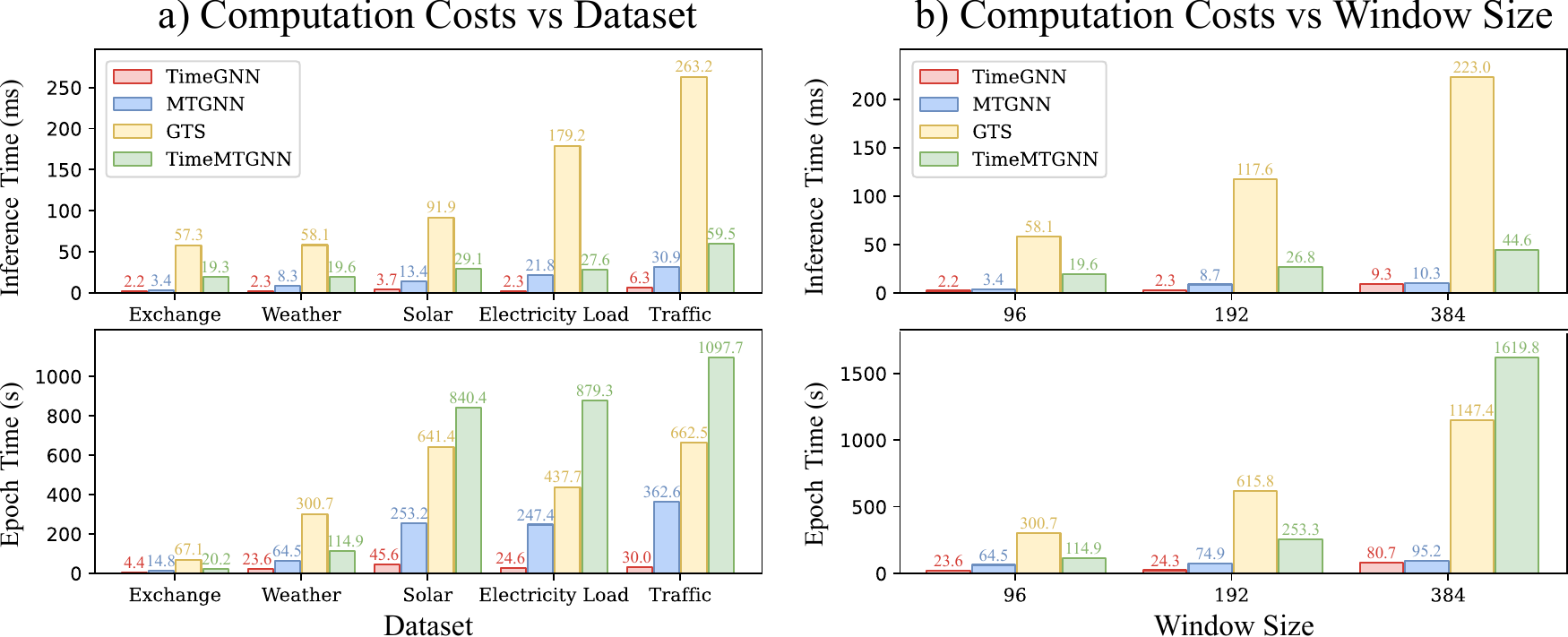}
    \caption{Computation costs of TimeGNN, TimeMTGNN and baseline models. (a) The inference and epoch training time per epoch between datasets. (b) The inference and epoch times with varying window sizes on the weather dataset}
    \label{fig:lambdas}
\end{figure}


\subsection{Experimental Setup} \label{sec:4.3}
Each model is trained for two runs for 50 epochs and the average mean squared error (MSE) and mean absolute error (MAE) score on the test set are recorded. The model chosen for evaluation is the one that performs the best on the validation set during training. 
The same dataloader is used for all models where the train, validation, and test splits are $0.7$, $0.1$, and $0.2$ respectively. The data is split first and each split is scaled using the standard scalar. The dataloader uses windows of length $96$ and batch size $16$. The forecasting horizons tested are $1$, $3$, $6$, and $9$ time steps into the future, where the exact value of the time step is dependent on the dataset (\eg $3$ time steps would correspond to $3$ hours into the future for the weather dataset and 3 days into the future for the Exchange dataset). In this paper, we use single-step forecasting for ease of comparison with other baseline methods. 
For training, we use the Adam optimizer with a learning rate of $0.001$.
Experiments for the Weather and Exchange datasets were conducted on an NVIDIA T4 and Electricity-Load, Solar, and Traffic on an NVIDIA A40. 
\begin{table}[h!]
    \caption{Forecasting performance for all multivariate datasets and baselines for different horizons $h$ - best in bold, second best underlined.}
    \label{tab:exchange-multi}
    \centering
    \tiny
    \def\arraystretch{1.3}
    \resizebox{\textwidth}{!}{
    \begin{tabular}{ |c c|c|c|c|c|c|c|c| }\hline
    \multicolumn{9}{|c|}{\multirow{2}{*}{\textbf{\scriptsize Exchange-Rate}}}\\
    & \multicolumn{7}{c}{} &\\
    \hline
       & \textbf{Metric} & \textbf{LSTM} & \textbf{TCN} & \textbf{LSTN} & \textbf{GTS} & \textbf{MTGNN} & \textbf{TimeGNN} & \textbf{TimeMTGNN}\\
      \hline
      \multirow{2}{*}{\rotatebox[origin=c]{90}{h=1}} 
        & mse & 0.328 ± 0.007 & 0.094 ± 0.118 & \textbf{0.004 ± 0.000} & \underline{0.005 ± 0.001} & 0.006 ± 0.002 & 0.129 ± 0.012 & \textbf{0.004 ± 0.001}\\ 
        & mae & 0.475 ± 0.033 & 0.191 ± 0.163 & \textbf{0.033 ± 0.000} & 0.041 ± 0.004 & 0.048 ± 0.011 & 0.294 ± 0.029 & \underline{0.034 ± 0.005}\\ 
      \hline
      \multirow{2}{*}{\rotatebox[origin=c]{90}{h=3}} 
        & mse & 0.611 ± 0.001 & 0.063 ± 0.035 & 0.013 ± 0.003 & \underline{0.009 ± 0.000} & 0.012 ± 0.000 & 0.368 ± 0.059 & \textbf{0.008 ± 0.001}\\ 
        & mae & 0.631 ± 0.031 & 0.190 ± 0.041 & 0.078 ± 0.012 & \underline{0.063 ± 0.000} & 0.078 ± 0.000 & 0.501 ± 0.045 & \textbf{0.061 ± 0.003}\\ 
      \hline
      \multirow{2}{*}{\rotatebox[origin=c]{90}{h=6}} 
        & mse & 0.877 ± 0.105 & 0.189 ± 0.221 & 0.033 ± 0.005 & \textbf{0.014 ± 0.001} & 0.024 ± 0.001 & 0.354 ± 0.031 & \underline{0.019 ± 0.004}\\ 
        & mae & 0.775 ± 0.032 & 0.290 ± 0.214 & 0.139 ± 0.008 & \textbf{0.081 ± 0.005} & 0.111 ± 0.000 & 0.453 ± 0.052 &\underline{0.099 ± 0.016}\\ 
      \hline 
      \multirow{2}{*}{\rotatebox[origin=c]{90}{h=9}} 
        & mse & 0.823 ± 0.118 & 0.123 ± 0.030 & 0.030 ± 0.006 & \textbf{0.020 ± 0.001} & 0.035 ± 0.003 & 0.453 ± 0.149 & \underline{0.034 ± 0.002}\\ 
        & mae & 0.743 ± 0.080 & 0.277 ± 0.037 & 0.124 ± 0.011 & \textbf{0.096 ± 0.001} & 0.140 ± 0.008 & 0.543 ± 0.084 & \underline{0.139 ± 0.010}\\ 
      \hline
    \multicolumn{9}{|c|}{\multirow{2}{*}{\textbf{\scriptsize Weather}}}\\
    & \multicolumn{7}{c}{} &\\
    \hline
       & \textbf{Metric} & \textbf{LSTM} & \textbf{TCN} & \textbf{LSTN} & \textbf{GTS} & \textbf{MTGNN} & \textbf{TimeGNN} & \textbf{TimeMTGNN}\\
      \hline
      \multirow{2}{*}{\rotatebox[origin=c]{90}{h=1}} 
        & mse & \textbf{0.162 ± 0.001} & \underline{0.176 ± 0.006} & 0.193 ± 0.001 & 0.209 ± 0.003 & 0.232 ± 0.008 & 0.178 ± 0.001 & 0.182 ± 0.003\\ 
        & mae & 0.202 ± 0.003 & 0.220 ± 0.011 & 0.236 ± 0.002 & 0.213 ± 0.004 & 0.230 ± 0.002 & \textbf{0.185 ± 0.000} & \underline{0.186 ± 0.000}\\ 
       \hline
       \multirow{2}{*}{\rotatebox[origin=c]{90}{h=3}} 
        & mse & \textbf{0.221 ± 0.000} & \underline{0.232 ± 0.003} & 0.233 ± 0.001 & 0.320 ± 0.005 & 0.263 ± 0.003 & 0.234 ± 0.001 & 0.234 ± 0.002 \\ 
        & mae & 0.265 ± 0.000 & 0.275 ± 0.000 & 0.285 ± 0.000 & 0.320 ± 0.001 & 0.273 ± 0.000 & \textbf{0.249 ± 0.001} & \underline{0.251 ± 0.001}\\ 
       \hline
       \multirow{2}{*}{\rotatebox[origin=c]{90}{h=6}} 
        & mse & \underline{0.268 ± 0.004} & 0.274 ± 0.002 & \textbf{0.266 ± 0.001} & 0.374 ± 0.003 & 0.301 ± 0.003 & 0.287 ± 0.002 & 0.282 ± 0.007\\ 
        & mae & 0.320 ± 0.004 & 0.323 ± 0.001 & 0.321 ± 0.000 & 0.388 ± 0.002 & 0.311 ± 0.002 & \textbf{0.297 ± 0.001} & \underline{0.300 ± 0.003}\\ 
       \hline 
       \multirow{2}{*}{\rotatebox[origin=c]{90}{h=9}} 
        & mse & \underline{0.292 ± 0.007} & 0.307 ± 0.009 & \textbf{0.288 ± 0.000} & 0.399 ± 0.002 & 0.329 ± 0.001 & 0.316 ± 0.001 & 0.311 ± 0.002\\ 
        & mae & 0.342 ± 0.003 & 0.350 ± 0.005 & 0.345 ± 0.003 & 0.420 ± 0.004 & \underline{0.339 ± 0.004} & \textbf{0.331 ± 0.001} & \textbf{0.331 ± 0.001}\\ 
      \hline
    \multicolumn{9}{|c|}{\multirow{2}{*}{\textbf{\scriptsize Electricity-Load}}}\\
    & \multicolumn{7}{c}{} &\\
    \hline
       & \textbf{Metric} & \textbf{LSTM} & \textbf{TCN} & \textbf{LSTN} & \textbf{GTS} & \textbf{MTGNN} & \textbf{TimeGNN} & \textbf{TimeMTGNN} \\
      \hline
      \multirow{2}{*}{\rotatebox[origin=c]{90}{h=1}}
        & mse & 0.226 ± 0.002 & 0.267 ± 0.001 & 0.064 ± 0.001 & 0.135 ± 0.002 & \textbf{0.046 ± 0.000} & 0.211 ± 0.003 & \underline{0.047 ± 0.000}\\ 
        & mae & 0.323 ± 0.000 & 0.375 ± 0.002 & 0.167 ± 0.001 & 0.246 ± 0.001 & \textbf{0.131 ± 0.000} & 0.309 ± 0.001 & \underline{0.135 ± 0.000}\\ 

       \hline
       \multirow{2}{*}{\rotatebox[origin=c]{90}{h=3}} 
        & mse & 0.255 ± 0.001 & 0.329 ± 0.015 &\textbf{ 0.065 ± 0.001} & 0.303 ± 0.019 & 0.079 ± 0.001 & 0.179 ± 0.003 & \underline{0.077 ± 0.000}\\ 
        & mae & 0.339 ± 0.000 & 0.406 ± 0.013 & \textbf{0.163 ± 0.002} & 0.388 ± 0.019 & \underline{0.171 ± 0.000} & 0.320 ± 0.002 & 0.173 ± 0.000\\ 

       \hline
       \multirow{2}{*}{\rotatebox[origin=c]{90}{h=6}} 
        & mse & 0.253 ± 0.005 & 0.331 ± 0.010 & 0.125 ± 0.006 & 0.334 ± 0.000 & \textbf{0.097 ± 0.000} & 0.246 ± 0.004 & \underline{0.104 ± 0.015}\\ 
        & mae & 0.340 ± 0.006 & 0.408 ± 0.009 & 0.238 ± 0.005 & 0.413 ± 0.000 & \textbf{0.189 ± 0.001} & 0.332 ± 0.004 & \underline{0.200 ± 0.016}\\ 
       \hline 
       \multirow{2}{*}{\rotatebox[origin=c]{90}{h=9}}
        & mse & 0.271 ± 0.009 & 0.349 ± 0.022 & 0.144 ± 0.013 & 0.289 ± 0.021 & \underline{0.108 ± 0.002} & 0.258 ± 0.010 & \textbf{0.104 ± 0.001}\\ 
        & mae & 0.351 ± 0.003 & 0.410 ± 0.019 & 0.251 ± 0.013 & 0.368 ± 0.020 & \underline{0.198 ± 0.002} & 0.344 ± 0.007 & \textbf{0.196 ± 0.001}\\ 
      \hline
    \multicolumn{9}{|c|}{\multirow{2}{*}{\textbf{\scriptsize Solar-Energy}}}\\
    & \multicolumn{7}{c}{} &\\
    \hline
       & \textbf{Metric} & \textbf{LSTM} & \textbf{TCN} & \textbf{LSTN} & \textbf{GTS} & \textbf{MTGNN} & \textbf{TimeGNN} & \textbf{TimeMTGNN}\\
      \hline
      \multirow{2}{*}{\rotatebox[origin=c]{90}{h=1}} 
        & mse & 0.019 ± 0.000 & 0.012 ± 0.000 & \underline{0.007 ± 0.000} & 0.012 ± 0.001 & \textbf{0.006 ± 0.000} & 0.022 ± 0.000 & \textbf{0.006 ± 0.000}\\ 
        & mae & 0.064 ± 0.000 & 0.055 ± 0.001 & \underline{0.035 ± 0.000} & 0.046 ± 0.003 & \textbf{0.026 ± 0.000} & 0.059 ± 0.000 & \textbf{0.026 ± 0.000}\\ 
      \hline
      \multirow{2}{*}{\rotatebox[origin=c]{90}{h=3}} 
        & mse & 0.031 ± 0.000 & \underline{0.030 ± 0.001} & 0.026 ± 0.000 & 0.044 ± 0.001 & \textbf{0.022 ± 0.002} & 0.030 ± 0.000 & \textbf{0.022 ± 0.000}\\ 
        & mae & 0.086 ± 0.002 & 0.087 ± 0.004 & 0.080 ± 0.000 & 0.098 ± 0.003 & \textbf{0.058 ± 0.002} & \underline{0.071 ± 0.000} & \textbf{0.058 ± 0.000}\\ 
      \hline
      \multirow{2}{*}{\rotatebox[origin=c]{90}{h=6}} 
        & mse & 0.046 ± 0.001 & 0.050 ± 0.000 & 0.049 ± 0.004 & 0.103 ± 0.001 & \textbf{0.042 ± 0.000} & 0.044 ± 0.000 & \underline{0.043 ± 0.002}\\ 
        & mae & 0.108 ± 0.005 & 0.121 ± 0.005 & 0.125 ± 0.013 & 0.163 ± 0.001 & \textbf{0.086 ± 0.001} & 0.090 ± 0.000 & \underline{0.088 ± 0.004}\\ 
      \hline 
      \multirow{2}{*}{\rotatebox[origin=c]{90}{h=9}} 
        & mse & 0.067 ± 0.003 & 0.073 ± 0.001 & 0.068 ± 0.000 & 0.167 ± 0.003 & \textbf{0.055 ± 0.001} & \underline{0.060 ± 0.002} & \underline{0.060 ± 0.000}\\ 
        & mae & 0.138 ± 0.009 & 0.150 ± 0.005 & 0.154 ± 0.004 & 0.218 ± 0.006 & \textbf{0.101 ± 0.001} & \underline{0.109 ± 0.001} & 0.110 ± 0.000\\ 
      \hline
    \multicolumn{9}{|c|}{\multirow{2}{*}{\textbf{\scriptsize Traffic}}}\\
    & \multicolumn{7}{c}{} &\\
    \hline
       & \textbf{Metric} & \textbf{LSTM} & \textbf{TCN} & \textbf{LSTN} & \textbf{GTS} & \textbf{MTGNN} & \textbf{TimeGNN} & \textbf{TimeMTGNN}\\
      \hline
      \multirow{2}{*}{\rotatebox[origin=c]{90}{h=1}} 
        & mse & 0.558 ± 0.007 & 0.594 ± 0.091 & \underline{0.246 ± 0.002} & 0.520 ± 0.010 & \textbf{0.233 ± 0.003} & 0.567 ± 0.002 & 0.293 ± 0.026\\ 
        & mae & 0.296 ± 0.005 & 0.352 ± 0.025 & 0.203 ± 0.002 & 0.319 ± 0.013 & \textbf{0.157 ± 0.002} & 0.281 ± 0.000 & \underline{0.162 ± 0.001}\\ 
      \hline
      \multirow{2}{*}{\rotatebox[origin=c]{90}{h=3}} 
        & mse & 0.595 ± 0.014 & 0.615 ± 0.002 & \underline{0.447 ± 0.010} & 0.970 ± 0.027 & \textbf{0.438 ± 0.001} & 0.622 ± 0.006 & 0.465 ± 0.012\\ 
        & mae & 0.318 ± 0.007 & 0.363 ± 0.003 & 0.286 ± 0.009 & 0.456 ± 0.010 & \textbf{0.205 ± 0.000} & 0.306 ± 0.002 & \underline{0.218 ± 0.007}\\ 
      \hline
      \multirow{2}{*}{\rotatebox[origin=c]{90}{h=6}} 
        & mse & 0.603 ± 0.001 & 0.680 ± 0.021 & \underline{0.465 ± 0.005} & 0.938 ± 0.048 & \textbf{0.450 ± 0.009} & 0.623 ± 0.004 & 0.495 ± 0.012\\ 
        & mae & 0.321 ± 0.003 & 0.403 ± 0.013 & 0.288 ± 0.002 & 0.461 ± 0.023 & \textbf{0.213 ± 0.003} & 0.311 ± 0.007 & \underline{0.239 ± 0.001}\\ 
      \hline 
      \multirow{2}{*}{\rotatebox[origin=c]{90}{h=9}} 
        & mse & 0.614 ± 0.011 & 0.655 ± 0.017 & \textbf{0.467 ± 0.010} & 0.909 ± 0.024 & \underline{0.471 ± 0.000} & 0.622 ± 0.002 & 0.494 ± 0.000\\ 
        & mae & 0.329 ± 0.010 & 0.382 ± 0.014 & 0.290 ± 0.006 & 0.453 ± 0.016 & \textbf{0.220 ± 0.002} & 0.313 ± 0.002 & \underline{0.236 ± 0.005}\\ 
      \hline
    \end{tabular}}
\end{table}

\subsection{Results}

\noindent \textbf{Scalability.} We compare the inference and training times of the graph-based models TimeGNN, MTGNN, GTS in Figure \ref{fig:lambdas}. These figures also include recordings from the ablation study of the TimeMTGNN variant, which is described in the relevant paragraph below. 
Figure~\ref{fig:lambdas}(a) shows the computational costs on each dataset. 
Among the baseline models, GTS is the most costly in both inference and training time due to the use of the entire training dataset for graph construction. 
In contrast, MTGNN learns static node features and is subsequently more efficient. 
In inference time, as the number of variables increases there is a noticeable increase in inference time for MTGNN and GTS as their graph sizes also increase. 
TimeGNN's graph does not increase in size with the number of variables and consequently, the inference time scales well across datasets. 
The training epoch times follow the observations in inference time. 

Since the size of the graphs used by TimeGNN is based on window size, the cost of increasing the window size on the weather dataset is shown in Figure~\ref{fig:lambdas}(b). 
As the window size increases, so does the cost of inference and training for all models. As the graph learning modules for MTGNN and GTS do not interact with the window size, the increase in cost can primarily be attributed to their forecasting modules.  
MTGNN's inference times do not increase as dramatically as GTS's, implying a more robust forecasting module. 
As the window size increases, TimeGNN's inference and training cost growth is slower than the other methods and remains the fastest of the GNN methods. 
The time-based graph learning module does not become overly cumbersome as window sizes increase. 

\noindent \textbf{Forecasting Quality.}  
Table \ref{tab:exchange-multi} summarizes the forecasting performance of the baseline models and TimeGNN for different horizons $h \in \{1,3,6,9\}$. 

In general, GTS has the best forecasting performance on the smaller Exchange-Rate dataset. The use of the training data during graph construction may give GTS an advantage over the other methods on this dataset. TimeGNN however shows signs of overfitting during training and is unable to match the other two GNNs. On the Weather dataset, the purely recurrent methods perform the best in MSE score across all horizons. TimeGNN is competitive with the recurrent methods on these metrics and surpasses the recurrent models on MAE. This suggests TimeGNN is producing more significant outlier predictions than the recurrent methods and TimeGNN is the best performing GNN method.  

On the larger Electricity-Load, Solar-Energy, and Traffic datasets, in general, MTGNN is the top performer with LSTNet close behind. However, for larger horizons, TimeGNN performs better than GTS and competitively with LSTNet and the other recurrent models. 
This shows that time-domain graphs can successfully capture long-term dependencies within a dataset although TimeGNN struggles more with short-term predictions. This could also be attributed to the simplicity of TimeGNN's forecasting module compared to the other graph-based approaches. 

\noindent \textbf{Ablation Study.} To empirically examine the effects of the forecasting module and the representation power of the proposed graph construction module in TimeGNN, we conducted an ablation study where we replaced MTGNN's graph construction module with our own, so-called TimeMTGNN baseline. The remaining modules and the hyperparameters in TimeMTGNN are kept as similar as possible to MTGNN. TimeMTGNN shows comparable forecasting performance to MTGNN on the larger Electricity-Load, Solar-Energy, and Traffic datasets and higher performance on the smaller Exchange-Rate and Weather datasets. 
This shows the TimeGNN graph construction module is capable of learning meaningful graph representations that do not impede and in some cases improve forecasting quality. 
As seen in Fig. \ref{fig:lambdas}, the computational performance of TimeMTGNN suffers in comparison to MTGNN. A major contributing factor is the number of graphs produced. MTGNN learns a single graph for a dataset while TimeGNN produces one graph per window, accordingly, the number of GNN operations is greatly increased. However, the focus of this experiment was to confirm that the proposed temporal graph-learning module preserves or improves accuracy over static ones rather than to optimize efficiency.

\section{Conclusion}
We have presented a novel method of representing and dynamically generating graphs from raw time series. 
While conventional methods construct graphs based on the variables, we instead construct graphs such that each time step is a node. 
We use this method in TimeGNN, a model consisting of a graph construction module and a simple GNN-based forecasting module, and examine its performance against state-of-the-art neural networks. While TimeGNN's relative performance differs between datasets, this representation is clearly able to capture and learn the underlying properties of time series. Additionally, it is far faster and more scalable than existing graph methods as both the number of variables and the window size increase. 

\section*{Acknowledgements}
This work is supported by the Wallenberg AI, Autonomous Systems and Software Program (WASP) funded by the Knut and Alice Wallenberg Foundation and the IP Paris ``PhD theses in Artificial Intelligence (AI)'' funding programme by ANR. Computational resources were provided by the National Academic Infrastructure for Supercomputing in Sweden (NAISS) at Alvis partially funded by the Swedish Research Council through grant agreement no. 2022-06725.

\bibliography{camera_ready}

\begin{thebibliography}{10}
\providecommand{\url}[1]{\texttt{#1}}
\providecommand{\urlprefix}{URL }
\providecommand{\doi}[1]{https://doi.org/#1}

\bibitem{bai2018empirical}
Bai, S., Kolter, J.Z., Koltun, V.: An empirical evaluation of generic convolutional and recurrent networks for sequence modeling. arXiv:1803.01271  (2018)

\bibitem{box2015time}
Box, G.E., Jenkins, G.M., Reinsel, G.C., Ljung, G.M.: Time series analysis: forecasting and control. John Wiley \& Sons (2015)

\bibitem{cao2020spectral}
Cao, D., Wang, Y., Duan, J., Zhang, C., Zhu, X., Huang, C., Tong, Y., Xu, B., Bai, J., Tong, J., et~al.: Spectral temporal graph neural network for multivariate time-series forecasting. Advances in neural information processing systems  \textbf{33},  17766--17778 (2020)

\bibitem{chauhan2015anomaly}
Chauhan, S., Vig, L.: Anomaly detection in ecg time signals via deep long short-term memory networks. In: 2015 IEEE international conference on data science and advanced analytics (DSAA). pp.~1--7. IEEE (2015)

\bibitem{cho2014learning}
Cho, K., Van~Merri{\"e}nboer, B., Gulcehre, C., Bahdanau, D., Bougares, F., Schwenk, H., Bengio, Y.: Learning phrase representations using rnn encoder-decoder for statistical machine translation. arXiv preprint arXiv:1406.1078  (2014)

\bibitem{ding2015deep}
Ding, X., Zhang, Y., Liu, T., Duan, J.: Deep learning for event-driven stock prediction. In: Twenty-fourth international joint conference on artificial intelligence (2015)

\bibitem{donner2010recurrence}
Donner, R.V., Zou, Y., Donges, J.F., Marwan, N., Kurths, J.: Recurrence networks—a novel paradigm for nonlinear time series analysis. New Journal of Physics  \textbf{12}(3),  033025 (2010)

\bibitem{gilmer2017neural}
Gilmer, J., Schoenholz, S.S., Riley, P.F., Vinyals, O., Dahl, G.E.: Neural message passing for quantum chemistry. In: International conference on machine learning. pp. 1263--1272. PMLR (2017)

\bibitem{gligorijevic2021structure}
Gligorijevi{\'c}, V., Renfrew, P.D., Kosciolek, T., Leman, J.K., Berenberg, D., Vatanen, T., Chandler, C., Taylor, B.C., Fisk, I.M., Vlamakis, H., et~al.: Structure-based protein function prediction using graph convolutional networks. Nature communications  \textbf{12}(1), ~3168 (2021)

\bibitem{hamilton2020time}
Hamilton, J.D.: Time series analysis. Princeton university press (2020)

\bibitem{hamilton2017inductive}
Hamilton, W., Ying, Z., Leskovec, J.: Inductive representation learning on large graphs. Advances in neural information processing systems  \textbf{30} (2017)

\bibitem{hochreiter1997long}
Hochreiter, S., Schmidhuber, J.: Long short-term memory. Neural computation  \textbf{9}(8),  1735--1780 (1997)

\bibitem{jang2016categorical}
Jang, E., Gu, S., Poole, B.: Categorical reparameterization with gumbel-softmax. arXiv preprint arXiv:1611.01144  (2016)

\bibitem{kearnes2016molecular}
Kearnes, S., McCloskey, K., Berndl, M., Pande, V., Riley, P.: Molecular graph convolutions: moving beyond fingerprints. Journal of computer-aided molecular design  \textbf{30},  595--608 (2016)

\bibitem{kipf2018neural}
Kipf, T., Fetaya, E., Wang, K.C., Welling, M., Zemel, R.: Neural relational inference for interacting systems. In: International conference on machine learning. pp. 2688--2697. PMLR (2018)

\bibitem{kosma2022time}
Kosma, C., Nikolentzos, G., Xu, N., Vazirgiannis, M.: Time series forecasting models copy the past: How to mitigate. In: Artificial Neural Networks and Machine Learning--ICANN 2022: 31st International Conference on Artificial Neural Networks, Bristol, UK, September 6--9, 2022, Proceedings, Part I. pp. 366--378. Springer (2022)

\bibitem{lacasa2008time}
Lacasa, L., Luque, B., Ballesteros, F., Luque, J., Nuno, J.C.: From time series to complex networks: The visibility graph. Proceedings of the National Academy of Sciences  \textbf{105}(13),  4972--4975 (2008)

\bibitem{lai2018modeling}
Lai, G., Chang, W.C., Yang, Y., Liu, H.: Modeling long-and short-term temporal patterns with deep neural networks. In: The 41st international ACM SIGIR conference on research \& development in information retrieval. pp. 95--104 (2018)

\bibitem{le2022deep}
Le~Guen, V., Thome, N.: Deep time series forecasting with shape and temporal criteria. IEEE Transactions on Pattern Analysis and Machine Intelligence  \textbf{45}(1),  342--355 (2022)

\bibitem{lecun1998gradient}
LeCun, Y., Bottou, L., Bengio, Y., Haffner, P.: Gradient-based learning applied to document recognition. Proceedings of the IEEE  \textbf{86}(11),  2278--2324 (1998)

\bibitem{li2017diffusion}
Li, Y., Yu, R., Shahabi, C., Liu, Y.: Diffusion convolutional recurrent neural network: Data-driven traffic forecasting. arXiv preprint arXiv:1707.01926  (2017)

\bibitem{lim2021time}
Lim, B., Zohren, S.: Time-series forecasting with deep learning: a survey. Philosophical Transactions of the Royal Society A  \textbf{379}(2194),  20200209 (2021)

\bibitem{lin2013network}
Lin, M., Chen, Q., Yan, S.: Network in network. arXiv preprint arXiv:1312.4400  (2013)

\bibitem{oreshkin2019n}
Oreshkin, B.N., Carpov, D., Chapados, N., Bengio, Y.: N-beats: Neural basis expansion analysis for interpretable time series forecasting. arXiv preprint arXiv:1905.10437  (2019)

\bibitem{rousseau2013graph}
Rousseau, F., Vazirgiannis, M.: Graph-of-word and tw-idf: new approach to ad hoc ir. In: Proceedings of the 22nd ACM international conference on Information \& Knowledge Management. pp. 59--68 (2013)

\bibitem{rubanova2019latent}
Rubanova, Y., Chen, R.T., Duvenaud, D.K.: Latent ordinary differential equations for irregularly-sampled time series. Advances in neural information processing systems  \textbf{32} (2019)

\bibitem{rumelhart1986learning}
Rumelhart, D.E., Hinton, G.E., Williams, R.J.: Learning representations by back-propagating errors. nature  \textbf{323}(6088),  533--536 (1986)

\bibitem{seo2018structured}
Seo, Y., Defferrard, M., Vandergheynst, P., Bresson, X.: Structured sequence modeling with graph convolutional recurrent networks. In: Neural Information Processing: 25th International Conference, ICONIP 2018, Siem Reap, Cambodia, December 13-16, 2018, Proceedings, Part I 25. pp. 362--373. Springer (2018)

\bibitem{shang2021discrete}
Shang, C., Chen, J., Bi, J.: Discrete graph structure learning for forecasting multiple time series. arXiv preprint arXiv:2101.06861  (2021)

\bibitem{vaswani2017attention}
Vaswani, A., Shazeer, N., Parmar, N., Uszkoreit, J., Jones, L., Gomez, A.N., Kaiser, {\L}., Polosukhin, I.: Attention is all you need. Advances in neural information processing systems  \textbf{30} (2017)

\bibitem{wu2021autoformer}
Wu, H., Xu, J., Wang, J., Long, M.: Autoformer: Decomposition transformers with auto-correlation for long-term series forecasting. Advances in Neural Information Processing Systems  \textbf{34},  22419--22430 (2021)

\bibitem{wu2020connecting}
Wu, Z., Pan, S., Long, G., Jiang, J., Chang, X., Zhang, C.: Connecting the dots: Multivariate time series forecasting with graph neural networks. In: Proceedings of the 26th ACM SIGKDD international conference on knowledge discovery \& data mining. pp. 753--763 (2020)

\bibitem{yu2017spatio}
Yu, B., Yin, H., Zhu, Z.: Spatio-temporal graph convolutional networks: A deep learning framework for traffic forecasting. arXiv preprint arXiv:1709.04875  (2017)

\bibitem{zhao2019t}
Zhao, L., Song, Y., Zhang, C., Liu, Y., Wang, P., Lin, T., Deng, M., Li, H.: T-gcn: A temporal graph convolutional network for traffic prediction. IEEE transactions on intelligent transportation systems  \textbf{21}(9),  3848--3858 (2019)

\bibitem{zhou2021informer}
Zhou, H., Zhang, S., Peng, J., Zhang, S., Li, J., Xiong, H., Zhang, W.: Informer: Beyond efficient transformer for long sequence time-series forecasting. In: Proceedings of the AAAI conference on artificial intelligence. vol.~35, pp. 11106--11115 (2021)

\end{thebibliography}
\bibliographystyle{splncs04}

\end{document}